\documentclass[twocolumn]{article}
\usepackage[utf8]{inputenc}
\usepackage{geometry}
\usepackage{xcolor}
\usepackage{graphicx}
\usepackage{booktabs}
\usepackage{multirow}
\usepackage{amsmath}
\usepackage{makecell}
\usepackage{hyperref}
\usepackage{amsfonts}
\usepackage{soul}
\usepackage{multicol}
\usepackage{makecell}
\usepackage{enumitem}
\usepackage{booktabs}
\usepackage{comment}
\usepackage[table]{xcolor}
\usepackage[mathlines,switch]{lineno}
\usepackage[square,sort,numbers]{natbib}
\usepackage[linesnumbered,ruled]{algorithm2e}

\hypersetup{hidelinks}

\title{Machine Collective Intelligence for Explainable Scientific Discovery}
\author{Gyoung S. Na$^{1,2}$ and Chanyoung Park$^2$\\
$^1$Korea Research Institute of Chemical Technology (KRICT)\\
$^2$Korea Advanced Institute of Science and Technology (KAIST)\\
\texttt{ngs0@krict.re.kr, cy.park@kaist.ac.kr}}

\date{}

\begin{document}

\maketitle

\begin{abstract}
Deriving governing equations from empirical observations is a longstanding challenge in science. Although artificial intelligence (AI) has demonstrated substantial capabilities in function approximation, the discovery of explainable and extrapolatable equations remains a fundamental limitation of modern AI, posing a central bottleneck for AI-driven scientific discovery. Here, we present machine collective intelligence, a unified paradigm that integrates two fundamental yet distinct traditions in computational intelligence--symbolism and metaheuristics--to enable autonomous and evolutionary discovery of governing equations. It orchestrates multiple reasoning agents to evolve their symbolic hypotheses through coordinated generation, evaluation, critique, and consolidation, enabling scientific discovery beyond single-agent inference. Across scientific systems governed by deterministic, stochastic, or previously uncharacterized dynamics, machine collective intelligence autonomously recovered the underlying governing equations without relying on hand-crafted domain knowledge. Furthermore, the resulting equations reduced extrapolation error by up to six orders of magnitude relative to deep neural networks, while condensing 0.5-1 million model parameters into just 5–40 interpretable parameters. This study marks an important shift in AI toward the autonomous discovery of principled scientific equations.
\end{abstract}

\subsection*{Artificial intelligence for science}
Artificial intelligence (AI) powered by deep neural networks (DNNs) has demonstrated notable capabilities in capturing high-dimensional latent relations between input and output variables from ground-truth observations \cite{deep_learning,tf}. In particular, large language models (LLMs) \cite{llm} and diffusion models \cite{diffu_model} are increasingly reshaping the scientific discovery workflow by enabling autonomous scientific reasoning and inverse design in a wide range of scientific domains \cite{llm_sci_discov,diffu_model_sci_discov}. These reasoning and generative frameworks have delivered substantial successes in solving longstanding scientific challenges, such as protein structure prediction \cite{alphafold}, inverse material design \cite{mattergen}, and fluid dynamics simulation \cite{ml_fluid_dym}.


However, the derivation of explainable and extrapolatable function (or equations) remains a fundamental limitation of modern AI, because DNNs inherently operate as a black-box function approximator that yields non-explainable interpolation functions expressed by millions of parameters \cite{deep_learning,ml_exp}. Despite substantial advances in connectionism-based AI, which is implemented by DNNs, this limitation poses a central bottleneck for AI-driven scientific discovery \cite{ai_lim_nat_sci1,ai_lim_nat_sci2}.

Although structural modification and post-interpretation methods have been developed to improve the explainability of connectionism-based AI \cite{lime,attn_vis_exp}, current AI models still have fundamental limitations in the explainability due to the following two reasons: (1) Most existing methods for explainable AI still produce non-symbolic interpretations, which cause subsequent tasks to understand their interpretation results \cite{lime}. (2) They typically impose rigorous constraints on input spaces or model architectures to produce explainable model outputs \cite{attn_vis_exp}.

\subsection*{Symbolism for artificial intelligence}
Unlike connectionism, symbolism \cite{symbolism} aims to find logical knowledge expressed by logical symbols and to simulate logical reasoning process of human intelligence. Symbolic regression grounded in symbolism is a representative approach for deriving explainable and extrapolatable equations (or functions) from empirical data \cite{sym_reg_review}. Inherently, symbolic regression yields human-readable mathematical equations and has been widely studied in statistics and machine learning toward the complete explainability of AI \cite{sym_reg_review2,sym_reg_review_sci}.

GPlearn is a symbolic regression framework that employs genetic programming to find computational programs that best describe the input-output relations \cite{gplearn}. PySR leverages a multi-population evolutionary algorithm to optimize computational trees that represents the relations between the input and output variables \cite{pysr}. In addition to these conventional methods, several methods that leverage DNN-based reinforcement learning and transformer architecture have been proposed for robust symbolic regression \cite{sym_reg_rl,sym_reg_tf}.

However, despite the notable explainability of symbolic regression, connectionism-based AI still dominate the field of AI for science. This is mainly because existing symbolic regression methods lack the regression capability required to capture highly nonlinear input-output relations encountered in real-world scientific systems \cite{lim_sym_reg}. Although a recent LLM-based symbolic regression method achieved substantial improvements in regression accuracy \cite{llm_sr}, its applicability to real-world problems is still questionable because its regression accuracy severely depends on the pretrained knowledge of backbone LLMs. In this regard, for successful symbolic regression, it is indispensable to develop an approach that can explore search spaces beyond the reasoning boundaries of backbone LLMs.

\subsection*{Computational collective intelligence}
While connectionism- and symbolism-based approaches focus on implementing computational intelligence that learns high-dimensional latent relations and distributions from previous observations, metaheuristic primarily aims to implement computational intelligence that explores search spaces to discover novel knowledge beyond previous observations \cite{metah_review}. Specifically, metaheuristic instantiates high-level and general-purpose computational intelligence by organizing iterative exploration, adaptation, and trial-and-error optimization into a unified framework \cite{sa,pso,aco}.

In particular, population-based metaheuristics implement collective intelligence across multiple agents for coordinated exploration and optimization through knowledge propagation and accumulation schemes \cite{pso,abc}. These population-based metaheuristics have demonstrated notable capabilities in solving complex search and optimization problems in a wide range of research fields, such as mathematical optimization \cite{metahu_comb}, chemical engineering \cite{metah_inv_eng}, and materials science \cite{ga_mat_design}. The effectiveness of the population-based metaheuristics largely stems from their knowledge-propagation and accumulation schemes, allowing individual agents to share best experiences and thereby accelerating global convergence beyond predefined knowledge \cite{metahu_review}.

\begin{figure}
    \centering
    \includegraphics[width=0.5\textwidth]{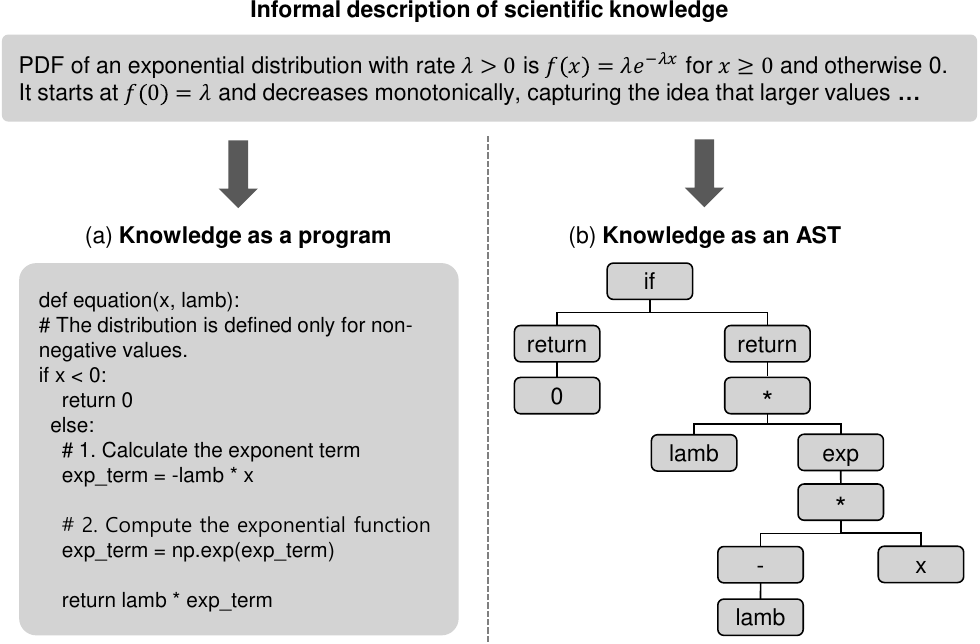}
    \vspace{-3ex}
    \caption{Conceptual comparison of two approaches for representing scientific knowledge. (a) Program- and (b) AST-based representations.}
    \label{fig:ast}
\end{figure}

\section*{Results}
\subsection*{Machine collective intelligence}
Here, we present Machine Collective Intelligence (MCI), a unified paradigm that integrates two fundamental yet distinct traditions in computational intelligence--symbolism and metaheuristics--for autonomous derivation of governing equations. MCI enables multiple reasoning agents to collectively discover symbolic equations for unknown scientific systems, without relying on hand-crafted programming commands and domain knowledge. To this end, MCI combines LLM-based reasoning agents with a population-based metaheuristic framework to facilitate knowledge accumulation and propagation across multiple reasoning agents, allowing scientific discovery beyond the prior knowledge of an individual reasoning agent.

\subsubsection*{Abstract syntax tree as a canonical representation of scientific knowledge}
Existing LLM-based symbolic regression methods typically formalize scientific knowledge as computer programs--sequences of logical symbols. In contrast, MCI formalizes scientific knowledge as an abstract syntax tree (AST) \cite{ast}, which is a structured formal representation of symbols and their logical flows. Fig. \ref{fig:ast} compares the program- and AST-based formal representations of scientific knowledge. In formal language theory, AST is used to represent canonical structures of computer programs. As illustrated in Fig. \ref{fig:ast}(b), AST reveals the essential logical flows of computer programs by abstracting away execution-irrelevant information, such as variable names and comments.

AST enables to quantify explainability of the discovered equations as a value inversely proportional to the tree depth of its corresponding AST; consequently, computer programs with lower logical complexity are prioritized as knowledge with high explainability. In the symbolic reasoning process, MCI calculates the explainabilities of the discovered equations and quantitatively compares them to identify the most compact equation among the derived equations.

\subsubsection*{Symbolic reasoning process}
\begin{figure*}
    \centering
    \includegraphics[width=1.0\textwidth]{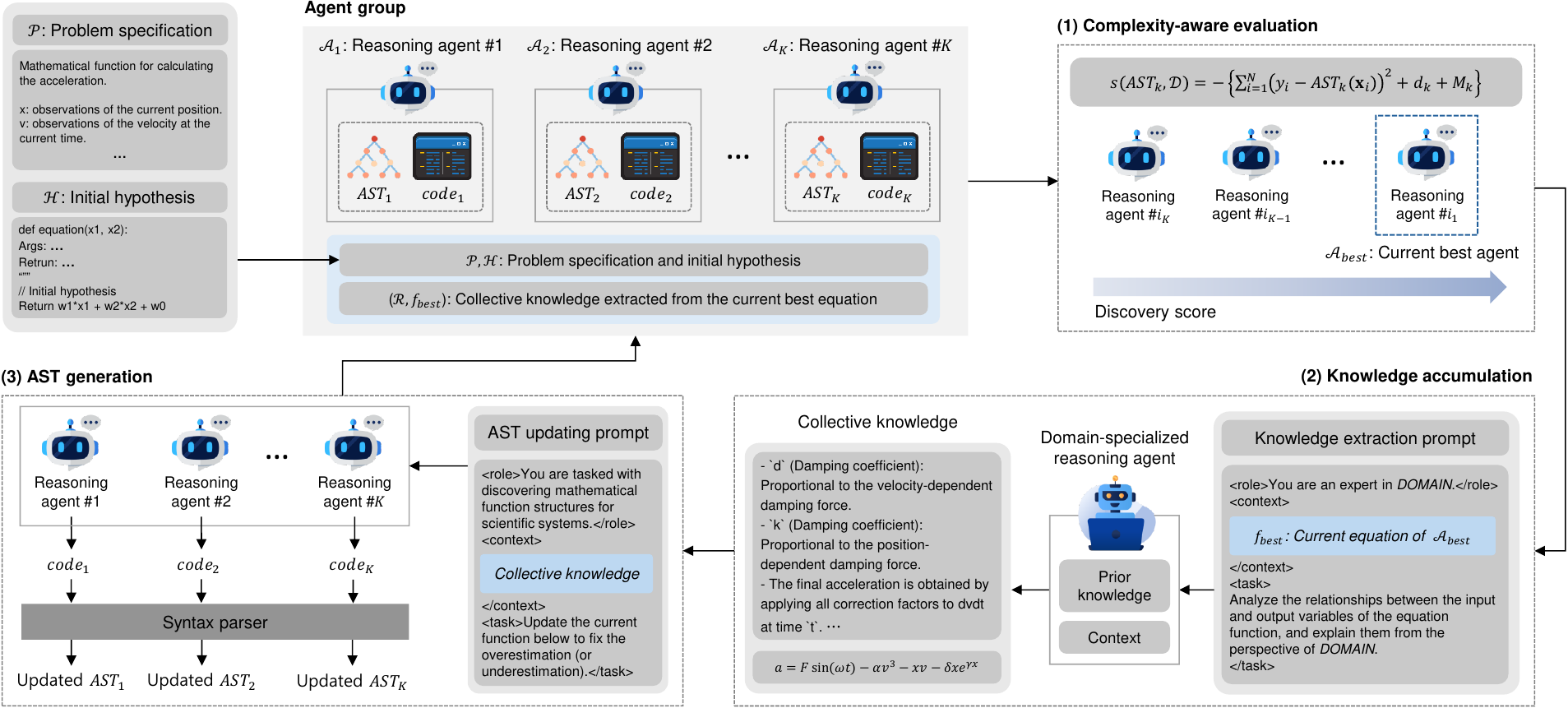}
    \vspace{-3ex}
    \caption{The overall symbolic reasoning process of MCI to discover governing equations for unknown scientific systems. MCI defines collective knowledge as a tuple of $f_{best}$ and its corresponding analysis result $\mathcal{R}$ in natural language.}
    \label{fig:mci}
\end{figure*}

Fig. \ref{fig:mci} illustrates the overall symbolic reasoning process of MCI to discover governing equations for unknown scientific systems. The symbolic reasoning process of MCI starts with an initialization step for experimental setup, followed by three iterative steps for scientific discovery.



\textbf{Step 0: initialization.} For a given hyperparameter $K \in \mathbb{N}^*$, MCI builds a group of LLM-based reasoning agents $\mathcal{A}_1, \mathcal{A}_2, ..., \mathcal{A}_K$, which are tasked with designing scientific equations. $\mathcal{A}_k$ has its own local memory that stores the current code and corresponding AST, denoted by $code_k$ and $AST_k$, respectively. After building the agent group, MCI registers user-defined problem specification $\mathcal{P}$ and initial hypothesis $\mathcal{H}$ in a group-level shared memory of the agent group. The problem specification describes input and output variables of the target scientific system, and initial hypothesis defines a frame of the equation written by programming languages. Given $\mathcal{P}$ and $\mathcal{H}$, the reasoning agents generate their own initial equations, after which a syntax parser converts the equations into ASTs.

\textbf{Step 1: complexity-aware evaluation.} In this step, MCI quantitatively evaluates the quality of the generated ASTs based on both their prediction errors and explainabilities. The prediction error of $AST_k$ is computed by the sum of squared errors (SSE) \cite{mse} over the set of ground-truth observations $\mathcal{D}$, i.e., $\sum_{(\textbf{x},y)\in\mathcal{D}}(y -AST_k(\textbf{x}))^2$, where $\mathbf{x}$ denotes a system input, $y$ is the corresponding ground-truth output, and $AST_k(\textbf{x})$ is the output of the equation canonicalized by $AST_k$. The explainability of $AST_k$ is quantified as the inverse of its depth, which represents logical complexity of the equation written by a programming language. Formally, MCI calculates the quality of $AST_k$ by Eq. \eqref{eq:ca_discov_score}, which we term the \textit{discovery score}.
\begin{equation} \label{eq:ca_discov_score}
    s(AST_k, \mathcal{D}) = -\left\{\sum_{(\textbf{x},y)\in\mathcal{D}}(y -AST_k(\textbf{x}))^2+d_k + M_k\right\},
\end{equation}
where $d_k$ is the depth of $AST_k$, and $M_k$ is the number of free parameters (learnable parameters) in $AST_k$. {In accordance with the minimum description length (MDL) principle \cite{mdl}, MCI minimizes the number of free parameters to distill the discovered equation into its most essential regularities, which enhances explainability by reducing the logical ambiguity from the free parameters.}

After calculating the discovery scores of $\mathcal{A}_1, \mathcal{A}_2, ..., \mathcal{A}_K$, MCI selects the reasoning agent achieving the highest score as the current best agent $\mathcal{A}_{best}$, and it is copied to the memory of MCI. The AST and its corresponding equation of $\mathcal{A}_{best}$ are subsequently designated as the current best AST and equation, denoted by $AST_{best}$ and $f_{best}$, respectively. Throughout the symbolic reasoning process, $AST_{best}$ serves as the group-level solution at the current iteration.

\textbf{Step 2: knowledge accumulation.} In this step, a reasoning agent specializing in a user-defined scientific domain conducts a stochastic reasoning process to analyze the logical structure of $f_{best}$ and interpret the physical meanings of its constituent variables. This produces an analysis result $\mathcal{R}$ that provides a natural-language explanation of $f_{best}$. MCI defines the \textit{collective knowledge} as the tuple $(\mathcal{R}, f_{best})$, which represents group-level insights extracted from $AST_{best}$. MCI deposits the collective knowledge into the shared memory within the agent group, and the reasoning agents leverages the collective knowledge when updating their own AST in the subsequent AST generation step, facilitating a collaborative reasoning process to discover the optimal governing equations for the target scientific system.

\textbf{Step 3: AST generation.} In this step, each reasoning agent generates new Python program implementing a candidate governing equation under the current local and global contexts, defined by the agent's current AST and the collective knowledge, respectively. For robust code generation, we designed a structured LLM prompt that embeds the local and global contexts alongside a code generation command. Detailed specifications of this structured prompt are provided in the Method section. After the code generation, the syntax parser converts the generated equations written in Python code into their corresponding ASTs. Subsequently, the symbolic reasoning process of MCI reverts to the complexity-aware evaluation (Step 1) and repeats the subsequent steps to iteratively refine the discovered governing equations.

\begin{algorithm}[ht!]
	\SetKwInOut{Input}{Input}
	\SetKwInOut{Output}{Output}
	\caption{The overall symbolic reasoning process of MCI for given scientific observations $\mathcal{D}$}
	\Input{$\mathcal{P}$: Problem specification,\\$\mathcal{H}$: Initial hypothesis,\\$\mathcal{Q}$: User-defined problem domain,\\$K \in \mathbb{N}$: Number of reasoning agents,\\$M \in \mathbb{N}$: Max iterations}
	\Output{$f_{best}$: the best equation at the final iteration}
	\label{alg:mci}
    $\mathcal{A} = \{\mathcal{A}_1, \mathcal{A}_2, ..., \mathcal{A}_K\}$ // Group of reasoning agents\\
    \BlankLine
    // Symbolic reasoning process for scientific discovery\\
    \For{$m=1;m \leq M;m=m+1$} {
        \For{$k=1;k \leq |\mathcal{A}|;k=k+1$} {
            \If{m == 1} {
                // For the initial code generation\\
                $code_k = \texttt{LLM}_{code}(\mathcal{P}, \mathcal{H})$
            }
            \Else {
                $code_k = \texttt{LLM}_{code}(AST_k, \mathcal{CK})$
            }
            $AST_k = \texttt{SyntaxParsing}(code_k)$\\
            $s_k = s(AST_k, \mathcal{D})$ // Eq. \eqref{eq:ca_discov_score}\\
        }
        $\mathcal{A}_{best} = \texttt{SelectBestAgent}(\mathcal{A}, \{s_1, s_2, ..., s_K\})$\\
        $f_{best} = \mathcal{A}_{best}.equation$\\
        $\mathcal{R} = \texttt{LLM}_{ans}(f_{best}, \mathcal{Q})$\\
        $\mathcal{CK} = (f_{best}, \mathcal{R})$
    }
    \Return{$f_{best}$}
\end{algorithm}

\begin{table*}[t]
    \centering
    \caption{Benchmark symbolic regression problems.}
    \vspace{1ex}
    \label{tb:problems}
    \small
    \begin{tabular}{c|l|lllcc}
    \toprule
    Domain
    & \makecell[l]{Problem Name\\(Target System)}
    & Input Variables
    & Output Variable
    & Governing Equation\\
    \toprule
    
    \multirow{4}{*}{\makecell{\\\\\\\\Physical\\science}}
    & \makecell[l]{\textbf{Chi2PDF} \cite{chi2_dist}\\(PDF of chi-squared\\distributions)}
    & \parbox{3.5cm}{
        \begin{itemize}[leftmargin=*, nosep]
            \item $x$: Data point
            \item $k$: Degree of freedom
        \end{itemize}
    }
    & \makecell[l]{Probability\\density}
    & $d_{x,k} = \frac{1}{2^{k/2}\Gamma(k/2)}x^{(k/2)-1}e^{-x/2}$\\
    \cline{2-6}
    
    & \makecell[l]{\textbf{NDO} \cite{llm_sr}\\(Nonlinear damped\\oscillator)}
    & \parbox{3.5cm}{
        \begin{itemize}[leftmargin=*, nosep]
            \item $t$: Time
            \item $x$: Position
            \item $v$: Velocity
        \end{itemize}
    }
    & Acceleration
    & $a = 0.3\sin(t) - 0.5v^3 - x v - 5x e^{0.5x}$\\
    \cline{2-6}

    & \makecell[l]{\textbf{NNN} \cite{deep_learning}\\(Nonlinear neural networks\\with sigmoid function)}
    & \parbox{3.5cm}{
        \begin{itemize}[leftmargin=*, nosep]
            \item $x_1$: 1st model input
            \item $x_2$: 2nd model input
        \end{itemize}
    }
    & Model output
    & \makecell[l]{Fully-connected neural network\\with two hidden layers in Eq. \eqref{eq:nnn}}\\
    \cline{2-6}

    & \makecell[l]{\textbf{MSB} \cite{msb}\\(Strain-stress relation\\in a metallic material)}
    & \parbox{3.5cm}{
        \begin{itemize}[leftmargin=*, nosep]
            \item $T$: Temperature
            \item $\epsilon$: Strain
        \end{itemize}
    }
    & \makecell[l]{Mechanical\\stress}
    & Unknown\\
    \hline
    
    \multirow{3}{*}{\makecell{\\\\\\\\Chemical\\science}}
    & \makecell[l]{\textbf{FHST} \cite{fhst}\\(Flory–Huggins equation\\for polymer mixing)}
    & \parbox{3.5cm}{
        \begin{itemize}[leftmargin=*, nosep]
            \item $\phi$: Volume fraction
            \item $N$: Chain length
            \item $\chi$: Interaction param.
        \end{itemize}
    }
    & \makecell[l]{Total change\\of Gibbs free\\energy}
    & \makecell[l]{$\Delta G_m = RT\left[\frac{\phi}{N}\ln\phi + \psi\ln\psi+\chi\phi\psi\right]$,\\where $R=8.314, T=300, \psi=1 - \phi$}\\
    \cline{2-6}

    & \makecell[l]{\textbf{BDC} \cite{bdc}\\(Lithium-ion battery)}
    & \makecell[l]{Six measurement\\parameters\\(cycle, voltage, etc.)}
    & \makecell[l]{Discharge\\capacity}
    & Unknown\\
    \cline{2-6}

    & \makecell[l]{\textbf{SFL} \cite{sfl}\\(Fatigue limit of alloy\\compositions)}
    & \makecell[l]{Weight Percents of\\ten elements and\\tempering temperature}
    & \makecell[l]{Fatigue limit}
    & Unknown\\
    \cline{2-6}

    & \makecell[l]{\textbf{NOMC} \cite{nomc}\\(Chemical reactor\\for non-oxidative\\methane conversion)}
    & \makecell[l]{Six engineering\\parameters\\(pressure, time, etc.)}
    & \makecell[l]{C$_2$ conversion\\yield}
    & Unknown\\
    \hline
    
    \multirow{2}{*}{\makecell{\\\\Biology}}
    & \makecell[l]{\textbf{ECBG} \cite{llm_sr}\\(Differential equation for\\bacterial growth rate of\\Escherichia coli)}
    & \parbox{3.5cm}{
        \begin{itemize}[leftmargin=*, nosep, before=\vspace{0.25\baselineskip}, after=\vspace{0.25\baselineskip}]
            \item $b$: Population density
            \item $s$: Substrate conc.
            \item $T$: Temperature
            \item $pH$: pH level
        \end{itemize}
    }
    & \makecell[l]{Bacterial\\growth rate}
    & \makecell[l]{Differential equation for\\bacterial growth rate in Eq. \eqref{eq:bactgrow}}\\
    \cline{2-6}
    & \makecell[l]{\textbf{HHM} \cite{hhm}\\(Differential equation for\\membrane potential in \\Hodgkin-Huxley model)}
    & \makecell[l]{Five biophysical\\parameters\\(membrane potential,\\gating probability, etc.)}
    & \makecell[l]{Rate of potential\\change}
    & \makecell[l]{Differential equation for\\Hodgkin–Huxley membrane\\potential dynamics in Eq. \eqref{eq:hhm}}
    &\\
    \bottomrule
    \end{tabular}
    \vspace{2ex}
\end{table*}

\subsubsection*{Overall symbolic reasoning process}
Algorithm \ref{alg:mci} formally describes the overall symbolic reasoning process of MCI to discover governing equations for unknown scientific systems. $\texttt{LLM}_{code}$ denotes a reasoning operation to generate a Python code representing the governing equation. $\texttt{LLM}_{ans}$ is an operation to analyze logical structures and physical interpretations of the generated Python code. \texttt{SyntaxParsing} is a function call to parse the Python code into its corresponding AST, and $\texttt{SelectBestAgent}$ is a operation to select the current best agent $\mathcal{A}_{best}$ according to the discovery scores $s_1, s_2, ..., s_K$. Finally, the algorithm returns $f_{best}$ and $\mathcal{CK}$ generated at the final iteration.

\begin{table*}[t]
    \centering
    \caption{In-distribution WMAPE values of MCI and the competitor methods on the benchmark symbolic regression problems. N/A indicates an execution failure caused by unresolved runtime errors.}
    \vspace{1ex}
    \label{tb:wmape_id}
    \begin{tabular}{c|cccc|c}
    \toprule
    Problem
    & GPlearn
    & PySR
    & LLM-SR
    & MSI
    & MCI\\
    \hline

    Chi2PDF
    & 0.6629 \scriptsize($\pm$0.0020)
    & 0.4656 \scriptsize($\pm$0.0184)
    & 0.0275 \scriptsize($\pm$0.0041)
    & 0.6915 \scriptsize($\pm$0.0838)
    & \bfseries2.22E-6 \scriptsize($\pm$3.13E-6)\\

    NDO
    & 0.4159 \scriptsize($\pm$0.0009)
    & 0.1713 \scriptsize($\pm$0.1741)
    & 0.0276 \scriptsize($\pm$0.0032)
    & 0.0642 \scriptsize($\pm$0.0062)
    & \bfseries1.01E-6 \scriptsize($\pm$2.84E-7)\\

    NNN
    & 0.5049 \scriptsize($\pm$0.0356)
    & N/A
    & 0.0048 \scriptsize($\pm$0.0017)
    & 0.0538 \scriptsize($\pm$0.0070)
    & \bfseries0.0015 \scriptsize($\pm$0.0001)\\

    MSB
    & 0.1962 \scriptsize($\pm$0.0206)
    & 0.1613 \scriptsize($\pm$0.0101)
    & 0.0401 \scriptsize($\pm$0.0045)
    & 0.0791 \scriptsize($\pm$0.0084)
    & \bfseries0.0281 \scriptsize($\pm$0.0033)\\

    FHST
    & 0.0820 \scriptsize($\pm$0.0202)
    & 0.0783 \scriptsize($\pm$0.0091)
    & 0.0117 \scriptsize($\pm$0.0010)
    & 0.0674 \scriptsize($\pm$0.0066)
    & \bfseries1.25E-7 \scriptsize($\pm$1.77E-7)\\
    
    BDC
    & 0.0989 \scriptsize($\pm$0.0044)
    & 0.0149 \scriptsize($\pm$0.0000)
    & 0.0154 \scriptsize($\pm$0.0038)
    & 0.0137 \scriptsize($\pm$0.0045)
    & \bfseries0.0092 \scriptsize($\pm$0.0011)\\

    SFL
    & 0.1844 \scriptsize($\pm$0.0097)
    & 0.1199 \scriptsize($\pm$0.0047)
    & 0.1131 \scriptsize($\pm$0.0202)
    & 0.0821 \scriptsize($\pm$0.0076)
    & \bfseries0.0432 \scriptsize($\pm$0.0012)\\
    
    NOMC
    & 0.2638 \scriptsize($\pm$0.0220)
    & 0.2031 \scriptsize($\pm$0.0001)
    & 0.1823 \scriptsize($\pm$0.0167)
    & 0.1658 \scriptsize($\pm$0.0147)
    & \bfseries0.0609 \scriptsize($\pm$0.0052)\\

    ECBG
    & 1.0087 \scriptsize($\pm$0.0079)
    & 1.4973 \scriptsize($\pm$0.0001)
    & 0.3182 \scriptsize($\pm$0.0875)
    & 0.3460 \scriptsize($\pm$0.0151)
    & \bfseries0.0742 \scriptsize($\pm$0.0079)\\

    HHM
    & 0.5949 \scriptsize($\pm$0.0048)
    & 0.8869 \scriptsize($\pm$0.0129)
    & 2.24E-7 \scriptsize($\pm$1.58E-7)
    & 0.0004 \scriptsize($\pm$0.0006)
    & \bfseries7.93E-8\scriptsize($\pm$4.79E-8)\\
    \bottomrule
    \end{tabular}
    \vspace{2ex}
\end{table*}

\begin{figure*}[t]
    \centering
    \includegraphics[width=0.99\textwidth]{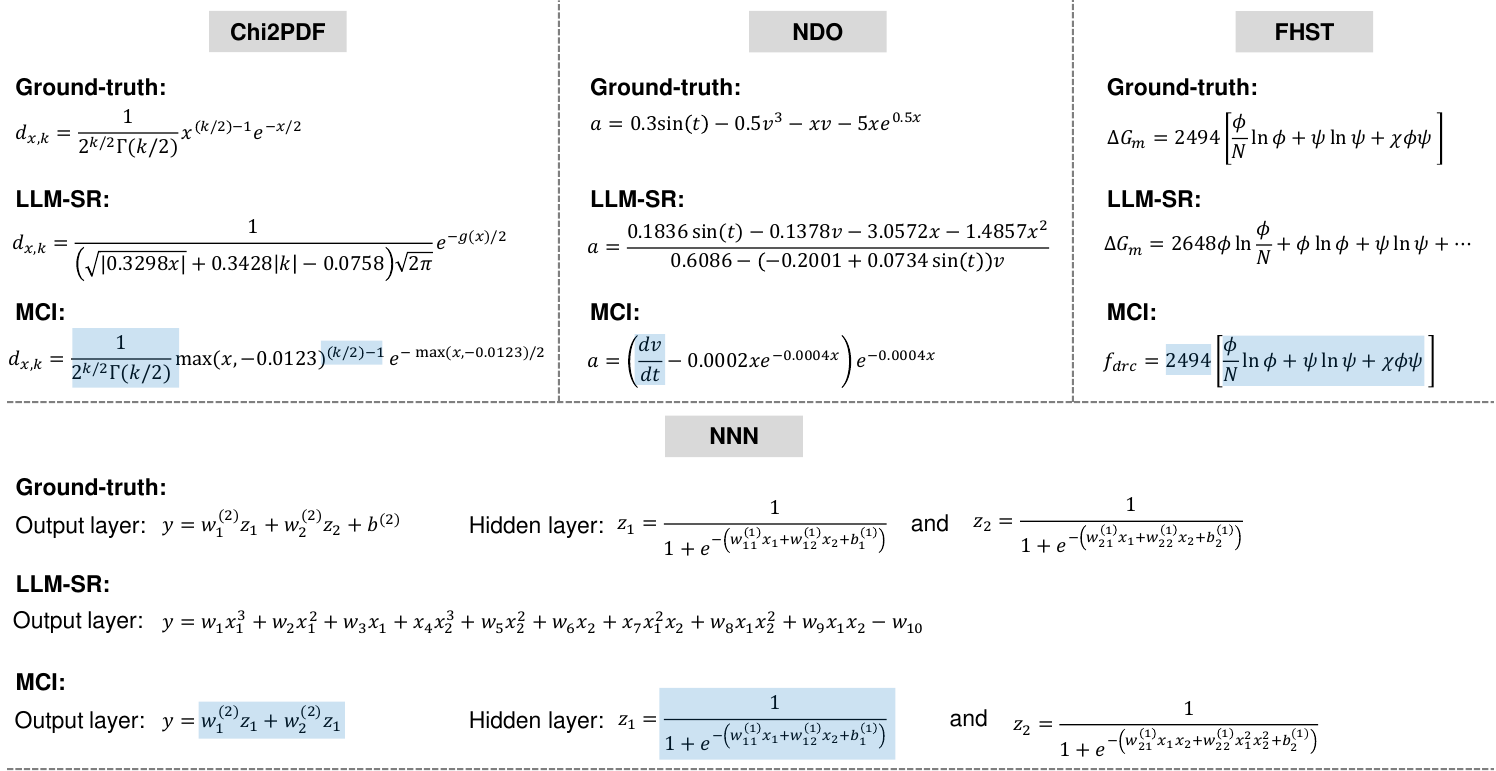}
    \caption{Ground-truth and discovered governing equations on the Chi2PDF, NDO, NNN, and FHST problems for which the governing equations are analytically given. $w_{ij}^{(k)}$ and $w_i^{(k)}$ are learnable weight parameters of the neural network.}
    \label{fig:quant_analysis}
\end{figure*}

\subsection*{Experimental evaluations}
\subsubsection*{Benchmark symbolic regression problems}
To rigorously assess the effectiveness of MCI in scientific discovery, we conducted experimental evaluations on benchmark symbolic regression problems across physics, chemistry, and biology. These benchmark problems were selected to cover a broad range of functional forms (e.g., polynomial, rational, transcendental, and composite expressions) and varying levels of uncertainty in governing equations. Table \ref{tb:problems} summarizes the problem specifications of the ten benchmark symbolic regression problems. Detailed descriptions of each benchmark are provided in the Method section.

\subsubsection*{Experimental settings}
We compared MCI against three state-of-the-art symbolic regression methods: GPlearn \cite{gplearn}, PySR \cite{pysr}, and LLM-SR \cite{llm_sr}. All competitor methods were configured according to their recommended settings, and hyperparameters were tuned within comparable computational budgets to ensure a fair comparison. In addition, we implemented an ablation variant termed Machine Single Intelligence (MSI) in which only a single reasoning agent performs evolutionary symbolic regression, without collective knowledge accumulation. Therefore, comparing MCI against MSI isolates the contribution of the collective intelligence of MCI, and this ablation study will demonstrate the efficacy of the collective intelligence on scientific discovery.

To ensure transparency and full reproducibility, MCI was developed independently of proprietary AI platforms (e.g., GPT-5 and Gemini). By avoiding dependence on closed-source AI systems, MCI facilitates broader access to its methodology and supports independent verification by the research community. To this end, all reasoning agents in MCI were implemented by Mixtral:8x7b \cite{mixtral}, which is an open-architecture LLM specialized for mathematical reasoning and programming.

In the experimental evaluations, the ground-truth observations were partitioned into two non-overlapping subsets: training and test datasets. The training dataset was used to calculate the discovery scores of the equations discovered by the symbolic regression methods. The test dataset was used exclusively to evaluate the generalization capability of the discovered equations.

The generalization error of the equations was quantified by weighted mean absolute percentage error (WMAPE) \cite{wmape} on the test dataset. WMAPE calculates the proportion of the prediction error relative to the total sum of ground-truth observation, providing a scale-invariant measure of percentage error. Formally, WMAPE is given by:
\begin{equation}
    \text{WMAPE} = \frac{\sum_{i=1}^N|y_i-\hat{y}_i|}{\sum_{i=1}^N|y_i|},
\end{equation}
where $y_i$ and $\hat{y}_i$ denote the ground-truth and model outputs, respectively. Lower WMAPE in the test dataset indicates better scale-invariant generalization accuracy. For a more rigorous evaluation, generalization errors quantified by normalized mean squared error (NMSE) \cite{nmse} and mean absolute error (MAE) \cite{mae} are additionally reported in the Supplementary Information. These supplementary results on diverse evaluation metrics confirms that the superiority of MCI is invariant to the choice of an evaluation metric.

\begin{figure*}[t]
    \centering
    \includegraphics[width=0.99\textwidth]{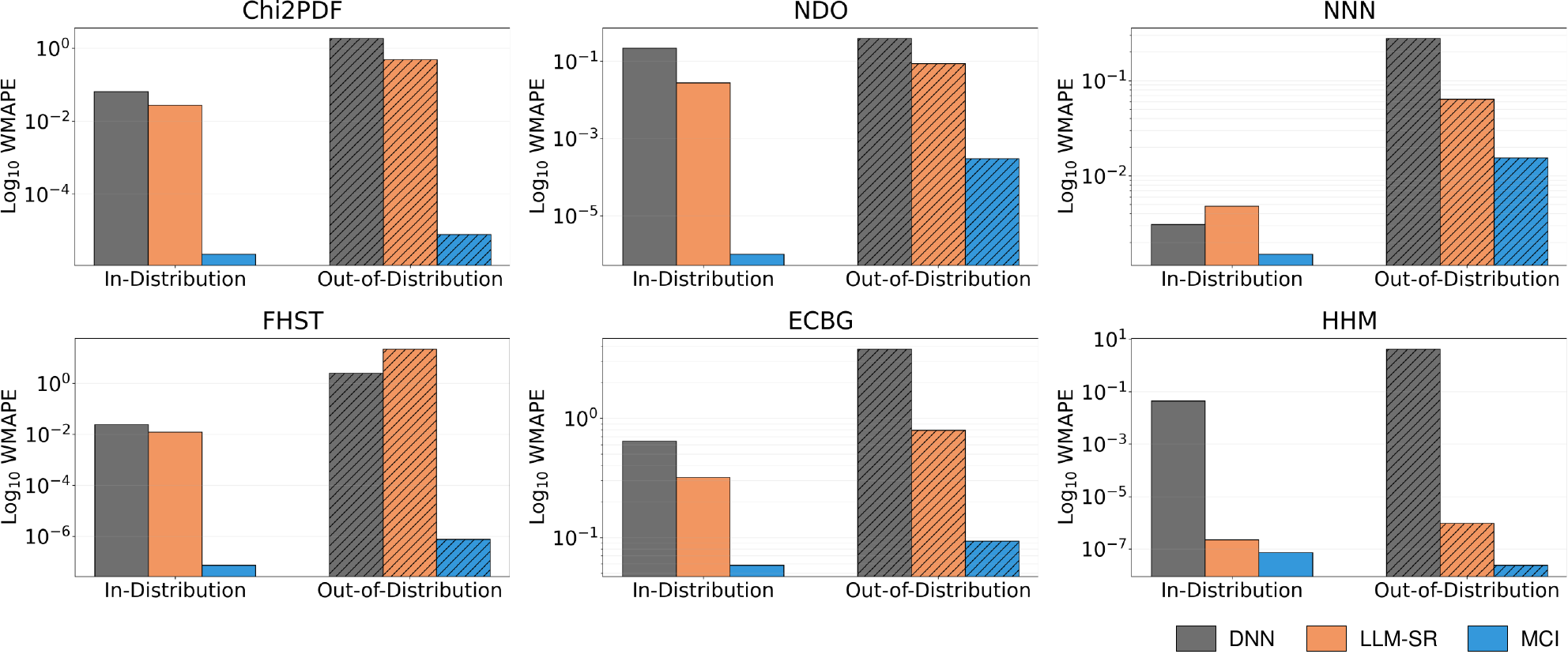}
    \caption{WMAPE of DNN, LLM-SR, and MCI under OOD conditions across six benchmark problems (Chi2PDF, NDO, NNN, FHST, ECBG, and HHM) for which the governing equations are known.}
    \label{fig:ood}
\end{figure*}

\subsubsection*{Symbolic regression accuracy}
We measured WMAPE of the equations discovered by the symbolic regression methods on the held-out test datasets generated under in-distribution conditions to assess their generalization capabilities to unseen scientific observations. All evaluations were repeated independently five times, and we reported the mean and standard deviation of WMAPE obtained from the repetitions. Table \ref{tb:wmape_id} presents WMAPE of the competitor symbolic regression methods and proposed MCI on the symbolic regression benchmarks.

In this experimental evaluation, we observed two primary results: First, the conventional symbolic regression methods, GPlearn and PySR, exhibited WMAPE values exceeding 0.1 on the Chi2PDF, NDO, MSB, SFL, NOMC, ECBG, and HHM problems, indicating an inability to accurately reconstruct the underlying governing equations. In particular, PySR failed to perform symbolic regression on the NNN problem due to unresolved runtime errors. Second, although LLM-SR achieved accuracy improvements over the conventional symbolic regression baselines, its WMAPE values still exceeded 0.1 on the SFL, NOMC, and ECBG problems. In contrast, MCI outperformed all symbolic regression methods and consistently yielded WMAPE below 0.1 across all benchmark problems. Quantitatively, MCI slashed 29.92-99.99\% generalization errors relative to those of the state-of-the-art method (LLM-SR), demonstrating its superior capacity for the autonomous derivation of scientific equations.

Fig. \ref{fig:quant_analysis} compares the ground-truth governing equations with the equations discovered by LLM-SR and MCI for the Chi2PDF, NDO, NNN, and FHST problems. To ensure a thorough analysis, we provided the original LLM-driven equations without any substitutions and simplifications in the Supplementary Information. In this evaluation, although LLM-SR also attained low WMAPE values on these problems, its scientific discovery process still largely relied on combinatorial enumerations over mathematical symbols to approximate input-output mappings, rather than uncovering the essential structures of the governing equations.

In contrast, MCI successfully uncovered the essential structures of the ground-truth governing equations on these benchmark problems. For the Chi2PDF problem, although MCI slightly mis-defined the domain of $x$ as $(-0.0123, +\infty)$ rather than the true range $(0, +\infty)$, it accurately captured the essential structure of the probability density function (PDF) of chi-squared distributions for arbitrary $x$ and the degrees of freedom $k$. On the NDO problem, MCI generated an equation expressed by the derivative of velocity $v$ with respect to time $t$ beyond an symbolic approximator to project the input variables to the output domain, indicating MCI discovered the underlying physics of the nonlinear damped oscillators behind the empirical observations. Finally, MCI perfectly reconstructed the governing equation on the FHST problem, despite the presence of multiple discontinuities and heuristically selected coefficients in the governing equation.


\begin{figure*}[t]
    \centering
    \includegraphics[width=0.99\textwidth]{fig/fig5.pdf}
    \vspace{-2ex}
    \caption{Prediction errors of the equations discovered by LLM-SR (black) and MCI (red). X-axis: Input variable of the equation. Y-axis: The absolute difference between ground-truth and predicted outputs. Gray-shaded area: In-distribution range of the input variable.}
    \label{fig:eq_ood_error}
\end{figure*}

\subsubsection*{Symbolic regression beyond reasoning boundaries of backbone LLMs}
The results on the NOMC problem are especially noteworthy because the target scientific system is our in-house chemical reactor with extensive chemical uncertainty. Since the pretrained knowledge of backbone LLMs does not fully encompass the underlying physics of our in-house chemical reactor, the NOMC problem provides a stringent benchmark for evaluating whether collective intelligence can discover scientific knowledge beyond the priors of a single reasoning agent.

As reported in Table \ref{tb:wmape_id}, LLM-SR showed a large WMAPE of 0.1823 on the NOMC problem, exhibiting a fundamental limitation of existing LLM-based symbolic regression methods: they excessively tend to rely on pretrained knowledge of backbone LLMs rather than exploring uncharted scientific knowledge. By contrast, MCI achieved a substantially lower WMAPE of 0.0609 and reduced 66.59\% error compared to the WMAPE of LLM-SR. This evaluation result on the NOMC problem demonstrates the potential of MCI for scientific discovery beyond prior knowledge of backbone LLMs.

\subsubsection*{Ablation study on collective intelligence}
To evaluate the efficacy of collective intelligence, we conducted an ablation study comparing in-distribution WMAPE values of MSI and MCI. As shown in Table \ref{tb:wmape_id}, despite sharing identical architectures and reasoning pipelines, MCI consistently outperformed MSI across all benchmark problems, achieving a significant reduction in generalization error of 39.42–99.99\%. Furthermore, the persistently higher standard deviations observed in MSI indicate that its final outcomes are inherently sensitive to initial seeding. This instability in MSI stems from the fact that its entire discovery process is contingent upon the initial hypotheses stochastically generated by a single reasoning agent. These results underscore the pivotal role of collective intelligence in enhancing the discovery capability of symbolic regression frameworks.

\subsubsection*{Numerical evaluation on OOD generalization}
Out-of-distribution (OOD) generalization remains a fundamental challenge for modern AI, including LLMs and generative diffusion models \cite{ml_challenges}. To assess extrapolation robustness, we generated ground-truth OOD samples for six benchmark problems for which the governing equations are known: Chi2PDF, NDO, NNN, FHST, ECBG, and HHM. Then, we compared WMAPE of the equations discovered by DNN, LLM-SR, and MCI on the OOD samples.

Fig. \ref{fig:ood} presents WMAPE of DNN, LLM-SR, and MCI on both in-distribution and OOD conditions. Although DNN achieved WMAPE below 0.1 for all benchmark problems, DNN failed to extrapolate on the Chi2PDF, FHST, ECBG, and HHM problems, where its WMAPE exceeded 1.0 on the OOD conditions. This result underscores the fundamental limitation of modern AI on extrapolation problems. LLM-SR likewise exhibited high WMAPE on the OOD samples, as its scientific discovery process still heavily relies on combinatorial enumerations to approximate input-output mappings as shown in Fig. \ref{fig:quant_analysis}, which lacks the OOD generalization.

By contrast, MCI maintained WMAPE $< 0.1$ for all benchmark problems. This robustness stems from i
ts emphasis on uncovering the essential structures of the governing equations beyond empirical input–output fitting, as described in Fig. \ref{fig:quant_analysis}. The primary distinction between LLM-SR and MCI lies in the implementation of collective intelligence, and the substantial improvement in OOD generalization achieved by MCI clearly demonstrates the efficacy of its collective intelligence in the scientific discovery. Collectively, these experimental results demonstrate MCI’s capacity to overcome the fundamental extrapolation challenges of modern AI.

\begin{figure}[t]
    \centering
    \includegraphics[width=0.48\textwidth]{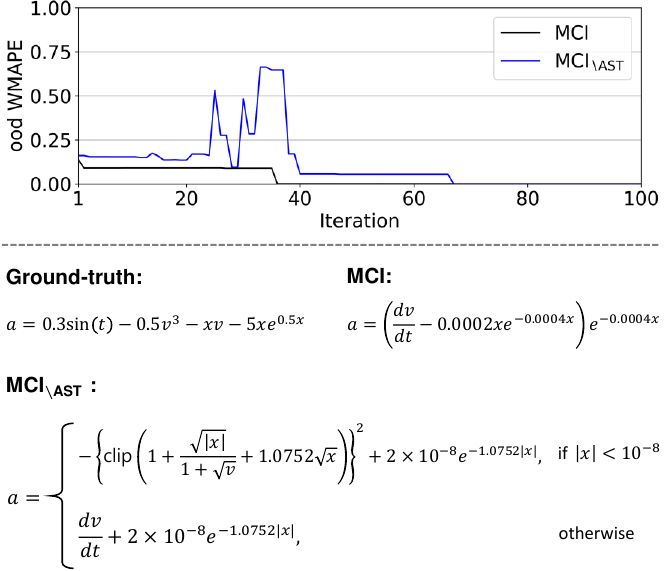}
    \vspace{-1ex}
    \caption{The governing equations derived by MCI and MCI$_{\setminus\text{AST}}$ with their respective OOD error convergence on the NDO problem. \texttt{clip} is a build-in function that constrains values to the range between 0 and 1.}
    \label{fig:ablation_ast}
\end{figure}

\subsubsection*{Qualitative analysis on OOD generalization}
Fig. \ref{fig:eq_ood_error} shows the absolute errors of the equations discovered by LLM-SR (black) and MCI (red) on the Chi2PDF, NDO, NNN, and FHST problems, for which the governing equations are analytically given. The absolute error is formally defined as the absolute difference between the ground-truth $y$ and predicted outputs $\hat{y}$, which is given by $|y - \hat{y}|$. For a comparative analysis, we presented a gray-shaded area to indicate the in-distribution range of the input variable.

Our experiments identified two critical limitations of the state-of-the-art symbolic regression method, LLM-SR, in OOD generalization: (1) Despite the averaged percentage errors below 0.1 on the NDO, NNN, and HMM problems, LLM-SR produced large absolute errors for many OOD input values. (2) While LLM-SR yields accurate predictions on in-distribution conditions for the NDO, NNN, ECGB, and HHM problems, its absolute errors escalated rapidly under OOD conditions, making a failure in extrapolation.

In contrast, MCI exhibited consistent generalization errors across the in-distribution and OOD conditions on the Chi2PDF, NDO, NNN, FHST, and ECGB problems. In addition, even when OOD inputs were generated from extreme ranges ($-100<V<-75$ and $35<I_{ext}<50$), MCI maintained the absolute error below $5\times10^{-4}$ in most cases, reducing 97.5\% WMAPE compared two it of LLM-SR under the OOD condition. Overall, the experiment results in Fig. \ref{fig:eq_ood_error} demonstrates the superior prediction accuracy and robustness of MCI under OOD conditions.

\subsubsection*{AST representation and explainability}
MCI employs AST as a formal representation of scientific knowledge, enabling the quantification of the explainability as shown in Eq. \eqref{eq:ca_discov_score}. To investigate the impact of the AST representation on symbolic reasoning, we compared the governing equations derived by MCI with those from its ablation variant lacking AST (MCI$_{\setminus\text{AST}}$). Fig. \ref{fig:ablation_ast} shows the governing equations derived by MCI and  MCI$_{\setminus\text{AST}}$ with their respective OOD error convergence on the NDO problem.

We observed two main results: (1) The complexity regularization by the AST representation facilitates faster and more robust OOD error convergence by prioritizing underlying physical laws over the localized function approximations, as shown in the OOD error convergence. (2) Although MCI$_{\setminus\text{AST}}$ captured the underlying physics expressed by the time derivative of $v$, it eventually yielded excessively complex equations. Consequently, there results presents that the AST representation and associated discovery score function are essential for deriving explainable and extrapolatable equations.

\section*{Discussion}
\subsection*{MCI and symbolic regression}
MCI distinguishes itself from conventional symbolic regression through three fundamental respects: (1) It replaces combinatorial enumeration with adaptive hypothesis evolution mediated by inter-agent critique and consolidation, enabling scientific discovery beyond the reasoning boundaries of individual intelligence. (2) It facilitates systematic knowledge accumulation and propagation for non-myopic exploration and progressive refinement of symbolic equations beyond a predefined symbol set. (3) It integrates a domain-specialized reasoning agent that extracts logical structures and physical meanings from candidate equations and feeds these insights back into the collective search as structured guidance.

\subsection*{Limitations and future work}
While this study employs an elitism-based knowledge propagation scheme \cite{elitism} to implement collective intelligence, there exist diverse implementation choices of alternative strategies for disseminating population-level knowledge among search agents \cite{elitism_method1, elitism_method2, elitism_method3}. Since well-tailored propagation schemes are known to improve both exploration and exploitation capabilities \cite{metahu_acc_review}, the symbolic reasoning performance of MCI might be further enhanced through alternative implementations of the knowledge propagation schemes. Consequently, comparative analyses of MCI for diverse propagation schemes remain a promising future research.

\bibliographystyle{naturemag}
\bibliography{ref}
\clearpage

\subsection*{Methods}
\subsubsection*{Structured LLM prompts for symbolic resoning}
\begin{figure}[t!]
    \centering
    \includegraphics[width=0.49\textwidth]{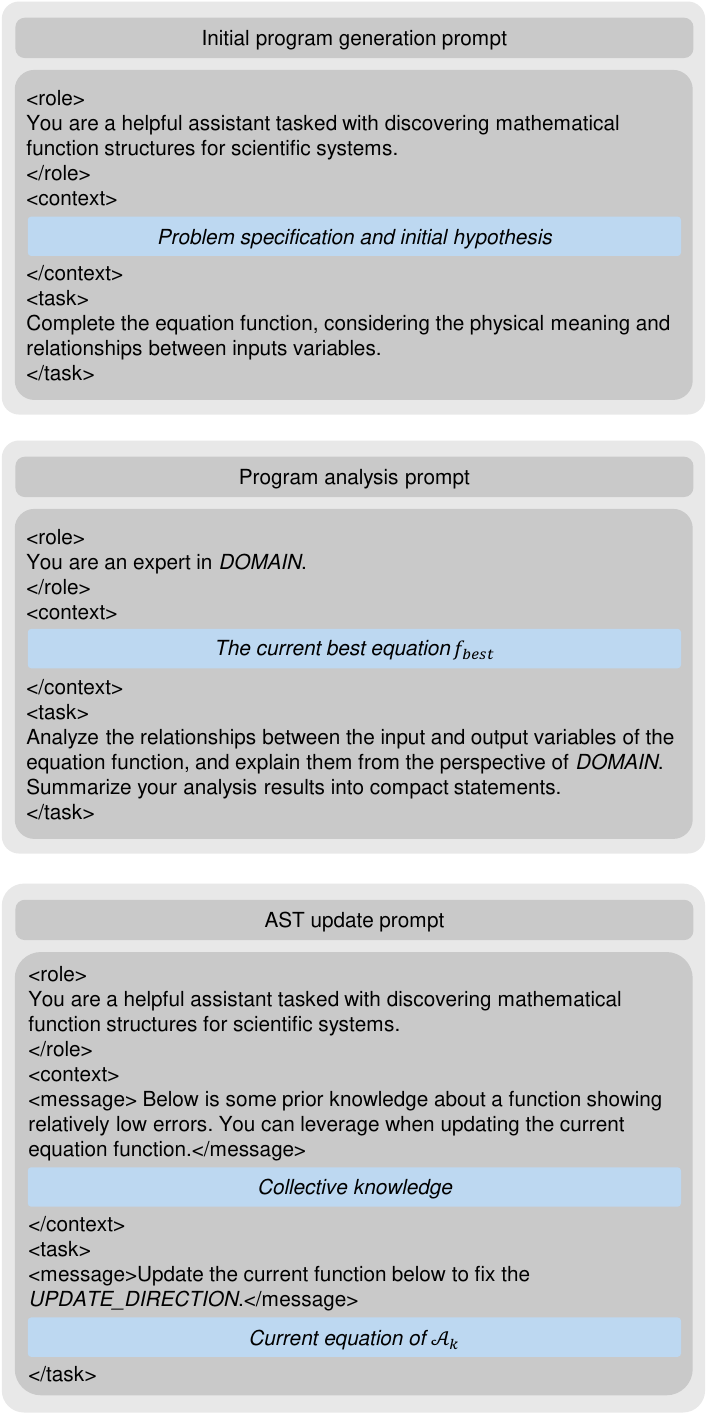}
    \vspace{-3ex}
    \caption{Structured LLM prompts to instruct the initial program generation, program analysis, and AST update in the symbolic reasoning process of MCI. \textit{DOMAIN} specifies a user-defined problem domain in the program analysis prompt, and \textit{UPDATE\_DIRECTION} is a variable assigned to either overestimation or underestimation based on the evaluation results of the current equation.}
    \label{fig:prompts}
\end{figure}

The symbolic reasoning process of MCI includes three LLM reasoning processes for: (1) initial program generation, (2) program analysis, and (3) AST update. Fig \ref{fig:prompts} presents structured LLM prompts used in each LLM reasoning process. The initial program generation prompt instructs the reasoning agents to a scientific equation written in Python from the problem specification and initial hypothesis. The program analysis prompt executes the domain-specialized reasoning agent to analyze logical structures and physical interpretations of the current best equation $f_{best}$, thereby generating collective knowledge across the reasoning agents. Finally, the AST update prompt directs the reasoning agents to revise their current equation based on the collective knowledge.

\subsubsection*{Implementation environment and model configuration}
We implemented MCI based on Python 3.12 and PyTorch 2.9.0 with CUDA 13.0. Mixtral:8x7b \cite{mixtral} was used as a backbone model of the reasoning agents in MCI. All implementations and experiments were conducted in a machine with Intel Xeon Gold 6526Y CPU, 64GB 5600MHz DDR5 memory, and NVIDIA RTX PRO 6000 Blackwell Max-Q GPU. Table \ref{tb:hparam_settings} presents the hyperparameter setting of MCI for each benchmark symbolic regression problem.

\subsubsection*{Specifications of the benchmark symbolic regression problems}

\small{\textbf{PDF of Chi-Squared Distributions (Chi2PDF).}} The chi-squared distribution with $k$ degree of freedom is the distribution of a sum of the squares of $k$ independent standard normal random variables \cite{chi2_dist}. It underpins a wide range of statistical procedures, including variance estimation and independence test. We defined a benchmark problem, Chi2PDF, which aims to recover the PDF of the chi-squared distribution for an arbitrary degree of freedom $k$. The ground-truth PDF of the chi-squared distribution at $x \geq 0$ is given by:
\begin{equation}
    f(x;k) = \frac{1}{2^{k/2}\Gamma(k/2)}x^{(k/2)-1}e^{-x/2},
\end{equation}
where $\Gamma$ is the gamma function. We randomly generated 1,000 $(x,k)$ pairs for symbolic regression and 200 non-overlapping $(x,k)$ pairs for an in-distribution evaluation.

\textbf{Nonlinear Damped Oscillator (NDO).} A nonlinear damped oscillator is an oscillatory system in which the restoring force and/or the damping force depends nonlinearly on displacement and velocity. Nonlinear damped oscillators frequently appears in a wide range of real-world scientific and engineering systems, such as aerospace and automotive vibration controllers, earthquake engineering, and biological self-oscillators. We defined the NDO problem by directly adopting the problem setting and ground-truth observations introduced in LLM-SR \cite{llm_sr}. The training and evaluation datasets each contain 10,000 ground-truth observations $(t, x, v, a)$, which represent time, position, velocity, and acceleration, respectively. The ground-truth acceleration was calculated through the governing equation below.
\begin{equation}
    a = F\sin(\omega t) - \alpha v^3 - \beta x v - \delta x \exp(\gamma x),
\end{equation}
where the constants $F, \omega, \alpha, \beta, \delta, \gamma$ were set to 0.3, 1.0, 0.5, 1.0, 5.0, 0.5, respectively.

\begin{table}[t]
    \centering
    \caption{The hyperparameter setting of MCI for each benchmark symbolic regression problem. The user-defined domain specifies the background knowledge of the domain-specialized reasoning agent in the knowledge accumulation step of MCI. $K$ and $M$ denote the number of reasoning agents and the maximum iterations, respectively.}
    \vspace{1ex}
    \label{tb:hparam_settings}
    \begin{tabular}{c|ccc}
    \toprule
    Problem
    & \makecell[c]{User-Defined Domain}
    & $K$
    & $M$\\
    \hline

    Chi2PDF
    & statistics
    & 50
    & 100\\

    NDO
    & physics
    & 50
    & 100\\

    NNN
    & machine learning
    & 50
    & 200\\
    
    MSB
    & materials science
    & 50
    & 100\\

    FHST
    & polymer science
    & 100
    & 50\\

    BDC
    & materials science
    & 50
    & 100\\

    SFL
    & materials science
    & 50
    & 100\\

    NOMC
    & organic chemistry
    & 50
    & 200\\

    ECBG
    & mathematical biology
    & 50
    & 200\\

    HHM
    & mathematical biology
    & 50
    & 100\\
    \bottomrule
    \end{tabular}
\end{table}

\textbf{Nonlinear Neural Network (NNN).} Neural networks is a fundamental building-block of current AI \cite{deep_learning}. The NNN problem aims to reveal a neural network architecture behind empirically observed model inputs and outputs. We implemented a fully-connected neural network with the sigmoid activation, which its output $y$ is formally given by:
\begin{equation} \label{eq:nnn}
    y = w_1^{(2)}z_1 + w_2^{(2)}z_2 + b^{(2)},
\end{equation}
where
\begin{equation}
    z_1 = \sigma(w_{11}^{(1)}x_1 + w_{12}^{(1)}x_2 + b_1^{(1)}),
\end{equation}
\begin{equation}
    z_2 = \sigma(w_{21}^{(1)}x_1 + w_{22}^{(1)}x_2 + b_2^{(1)}),
\end{equation}
and $[x_1, x_2]$ is a 2-dimensional input vector, $w_{ij}^{(k)}$ (or $w_i^{(2)})$ are learnable weights, $b_i^{(k)}$ (or $b^{(2)}$) are learnable biases, and $\sigma$ is the sigmoid function for nonlinear activation. We randomly generated 10,000 $(x_1, x_2, y)$ pairs for symbolic regression and 2,000 non-overlapping $(x_1, x_2, y)$ pairs for an in-distribution evaluation.

\textbf{Material Stress Behavior (MSB).} The MSB problem, introduced in LLM-SR, is a symbolic regression task that aims to infer the stress–strain behavior of Aluminium 6061-T651 at a specific temperature \cite{llm_sr,msb}. Given strain and temperature as inputs, symbolic regression methods are tasked with discovering a governing equation for the corresponding stress. The MSB problem has two inherent challenges: (1) the ground-truth observations are obtained from real chemical experiments and therefore contain measurement noise and experimental error. (2) there is no predetermined theoretical model to represent the stress-strain relationships, making the symbolic regression problem intrinsically hard for LLMs.

\textbf{Flory-Huggins solution theory (FHST).} Flory-Huggins solution theory \cite{fhst} characterizes the thermodynamic landscape of the polymer system based on a mean-field lattice model. The Gibbs free energy of mixing per lattice site, $\Delta G_m$, is defined as:
\begin{equation} \label{eq:fhst}
    \Delta G_m = RT\left[\frac{\phi}{N}\ln\phi + (1-\phi)\ln(1-\phi) + \chi\phi(1-\phi) \right],
\end{equation}
where $\phi$ denotes the volume fraction of the polymer, $N$ represents the chain length of the polymer, and $\chi$ is the dimensionless Flory-Huggins interaction parameter accounting for the enthalpy of mixing. $R$ and $T$ were manually determined as 8.314 and 300. This benchmark contains several singularity points and manually selected constants, yielding a fundamental challenge in symbolic regression. Based on the ground-truth equation in Eq. \eqref{eq:fhst}, we generated 10,000, 2,000, and 2,000 non-overlapping samples for training, in-distribution testing, and OOD evaluation, respectively.

\textbf{Battery discharge capacity (BDC).} To make this symbolic regression benchmark, we employed a well-known lithium-ion battery aging dataset extracted from a battery database provided by the NASA Ames Prognostics Center of Excellence (PCoE). A refined version of the dataset is available at \url{https://github.com/alg-x/Battery-Capacity-Prediction-Using-Regression}. The dataset contains experimental records of 18650-type Li-ion cell subjected to repetitive charge, discharge, and electrochemical impedance spectroscopy cycles at 24 Celsius. This dataset provide experimentally measured capacity for six measurement input variables: cycle index, battery voltage, battery current, battery temperature, voltage load, and current load. We used 334 and 83 instances for training and evaluating the symbolic regression methods, respectively.

\textbf{Steel fatigue limit (SFL).}
The steel fatigue strength dataset, sourced from the National Institute for Materials Science (NIMS) database, consists of 437 experimental records correlating manufacturing parameters with the fatigue limits of product alloys \cite{sfl}. We identified ten primary features for each record: the weight percentages (wt\%) of nine chemical elements (C, Si, Mn, P, S, Ni, Cr, Cu, and Mo) and the temperature of the final heat treatment process. For rigorous accuracy evaluation, the dataset was partitioned into a training dataset of 350 records and a non-overlapping evaluation dataset of 87 records.

\textbf{Non-Oxidative Methane Conversion (NOMC).} Non-oxidative methane conversion (NOMC) is a promising technology for mitigating greenhouse-gas emissions from industrial applications \cite{nomc_app}. We built an in-house NOMC reactor and measured conversion performance as a function of six experiment parameters: pressure, temperature, flow rate, H$_2$ feed amount, reactor length, and reactor diameter. We use 200 observations for symbolic regression and reserve a non-overlapping set of 51 observations for testing. The NOMC problem can be categorized as one of the most challenging symbolic regression problems due to three reasons: (1) the input-output relations are strongly nonlinear and contains inherent uncertainty from chemical reactions. (2) theoretical models to calculate the NONC performances of chemical reactors has not been established. (3) Because the ground-truth observations were obtained from our in-house chemical reactor, the backbone LLM is unlikely to possess sufficient prior knowledge of the target system, thereby imposing the challenging requirement of scientific discovery beyond its pretrained knowledge.

\textbf{Escherichia Coli Bacterial Growth (ECBG).} Quantifying the growth dynamics of Escherichia coli bacteria is a cornerstone of bacterial toxicity studies and food safety analytics. Typically, the bacterial population growth rate has been modeled using nonlinear differential equations. The ECBG problem, introduced in LLM-SR, aims to discover a nonlinear differential equation governing the bacterial growth rate. The ground-truth governing equation is given by:
\begin{equation} \label{eq:bactgrow}
    \frac{dB}{dt} = \mu_{max}B\left(\frac{S}{K_s + S}\right)\frac{\tanh(k(T - x_0))}{1 + c(T - x_{decay})^4}f_{pH},
\end{equation}
where the pH dependency $f_{pH}$ is given by:
\begin{equation}
    f_{pH} = \exp(-|pH - pH_{opt}|)\sin\left(\frac{(pH - pH_{min})\pi}{pH_{max} - pH_{min}}\right)^2,
\end{equation}
$B$ denotes bacterial population density, $S$ is substrate concentration, $T$ is temperature, and pH is the acidity or alkalinity of the growth medium. We employed the training and test datasets from LLM-SR, which contain 7,500 and 2,500 observations, respectively. The full implementation details of the ground-truth differential equation are described in the original LLM-SR paper \cite{llm_sr}.

\textbf{Hodgkin-Huxley model (HHM).} The Hodgkin-Huxley model is a foundational conductance-based framework to represent the electrophysiological dynamics of the system \cite{hhm}. This model describes the generation and propagation of action potentials through a set of coupled non-linear ordinary differential equations. The lipid bilayer is treated as a parallel capacitor, where the total membrane current is the sum of the capacitive current and ionic currents:
\begin{equation} \small \label{eq:hhm}
    C_m\frac{dV}{dt} = \bar{g}_{Na}m^3h(V_{Na} - V) + \bar{g}_Kn^4(V_K - V) + \bar{g}_L(V_L-V) + I_{ext},
\end{equation}
where $V$ denotes membrane potential, $m$ is sodium activation gating probability, $n$ is sodium inactivation gating probability, $h$ is potassium activation gating probability, and $I_{ext}$ represents external current. $C_m$ was fixed to 1.0, and other constants $(\bar{g}_{Na}, \bar{g}_{K},\bar{g}_L,V_{Na},V_{K},V_{L})$ are determined manually within physically reasonable ranges. Based on the ground-truth equation in Eq. \eqref{eq:hhm}, we generated 10,000, 2,000, and 2,000 non-overlapping samples for training, in-distribution testing, and OOD evaluation, respectively.

\subsection*{Data availability}
The training and evaluation datasets of the Chi2PDF, NNN, FHST, NOMC, and HHM problems are available at \url{https://github.com/ngs00/mci}. Those of the NDO, MSB, and ECGB problems are available at \url{https://github.com/deep-symbolic-mathematics/LLM-SR}. The original problem names of NDO, MSB, and ECBG in the LLM-SR repository are oscillator2, stressstrain, and bactgrow, respectively. The original data source of the BDC problem is \url{https://github.com/alg-x/Battery-Capacity-Prediction-Using-Regression}. The original dataset of the SFL problem is available at \url{https://link.springer.com/article/10.1186/2193-9772-3-8#MOESM1}.

\subsection*{Code availability}
The source code of MCI and experiment scripts of this paper are publicly available at \url{https://github.com/ngs00/mci}.

\end{document}